# Too good to be true? Predicting author profiles from abusive language

Isabelle van der Vegt[1], Bennett Kleinberg[1,2] and Paul Gill[1]

[1] Department of Security and Crime Science, University College London

[2] Dawes Centre for Future Crime, University College London

{isabelle.vandervegt, bennett.kleinberg, paul.gill}@ucl.ac.uk

## Abstract

The problem of online threats and abuse could potentially be mitigated with a computational approach, where sources of abuse are better understood or identified through author profiling. However, abusive language constitutes a specific domain of language for which it has not yet been tested whether differences emerge based on a text author's personality, age, or gender. This study examines statistical relationships between author demographics and abusive vs normal language, and performs prediction experiments for personality, age, and gender. Although some statistical relationships were established between author characteristics and language use, these patterns did not translate to high prediction performance. Personality traits were predicted within 15% of their actual value, age was predicted with an error margin of 10 years, and gender was classified correctly in 70% of the cases. These results are poor when compared to previous research on author profiling, therefore we urge caution in applying this within the context of abusive language and threat assessment.

## 1 Introduction

In June 2016, a far-right terrorist murdered Labour MP Jo Cox during the United Kingdom's EU referendum campaign [1]. Prior to the UK elections in December 2019, a record number of female MPs stood down citing the constant abuse and threats they endure [2]. Violent threats to politicians and public figures remain a serious problem, in particular due to the rise of threats communicated over the internet. Computational linguistics can potentially play a key role in better understanding and mitigating this social phenomenon.

In recent years, increased efforts focused on understanding and detecting abusive language and hate speech. These endeavours may be of particular interest to law enforcement and tech companies mitigating online threats, who wish to increase insight and reduce human workload. Studies examined abusive posts on social media, comment sections and forums [3, 4], online extremist language use [5, 6] and the development of bespoke threat assessment tools [7]. With this line of research, studies often refer to an established link between language and personality, [8] age, [9] and gender [10]. A closer examination of these latter studies, however, demonstrates the majority obtained small effects [11, 12] and the accuracy varies widely when predicting author characteristics [13-16].

This raises two questions. First, whether the link between language and author characteristics translates to contexts focused upon threats and abuse. Second, whether the small, yet statistically significant, relationships between language and author demographics can be adequately translated into prediction systems for practice.

The current study presents an experiment in which participants write a neutral, non-offensive text, and an abusive text directed at a politician. In line with abovementioned questions, our aim in this paper is to 1) examine the relationships between author characteristics (personality, age, and gender) and language, with a special focus on abusive language and 2) predict author profiles based on the linguistic characteristics of texts, where our novel contribution is the application to abusive texts.

## 1.1 Statistical relationships between language and author characteristics

Early studies using automated approaches to studying language departed from the assumption that linguistic style differs between individuals [8]. Specific traits such as the Big Five personality traits (Openness, Conscientiousness, Extraversion, Agreeableness, and Neuroticism) were correlated with certain linguistic characteristics, such as the use of negative emotion words, negations, and present tense [8]. Language use is frequently measured with the Linguistic Inquiry and Word Count (LIWC) software [9, 10], which measures the proportions of words from categories representing linguistic dimensions and grammar (e.g. personal pronouns, verbs), psychological processes (e.g. positive emotion, insight, hearing), and personal concerns (e.g. money, religion). The LIWC has been applied to a sample of psychology students' writing ($n$ = 1203), who wrote a 'stream of consciousness' essay describing current thoughts, feelings, and sensations. Results showed small positive correlations between neuroticism and negative emotion words ($r$=0.16), and a positive correlation between positive emotion words ($r$=0.15), social references ($r$=0.12) and extraversion [8]. Other endeavours [11] also showed correlations between personality traits and LIWC categories ($r$=0.23).

Research also examines age effects on language use. A large-scale study observed older age associated with decreased references to the self and others, and increased use of present- and future-tense over past-tense verbs [12]. Increased age was also associated with an increase in positive emotion words ($r$=0.05) and a decrease in negative emotion words ($r$=-0.04) [12]. Gender differences in language emerged in a study of 14,324 text samples including stream of consciousness essays [13]. Women more often used LIWC categories such as pronouns (Cohen's $d$=0.36[1]) and social words ($d$=0.21).

## 1.2 Predicting author characteristics from language

Linguistic information has also been used to predict personality traits, age, and gender. For personality prediction, ground truth for supervised learning tasks is often obtained directly through administering personality scales. Other approaches may label texts through third-person annotation or other personality correlates. In one example, participants completed a personality survey and wrote stream-of-consciousness essays, after which neuroticism and extraversion were predicted [15]. A binary classification task classified participants as either high (top third) or low (bottom third) scorers on the traits. Various psycholinguistic measures were used as features, and the average classification accuracy was 58% [15]. In a similar effort, $n$-grams were used as features to predict Big Five scores in several binary and multiclass prediction tasks [16]. Accuracies ranged from 45% (e.g., for five-class extraversion; random baseline: 33.8%) to 100% (e.g., for binary agreeableness, baseline: 54.2%) depending on the task, personality trait and feature set [16].

Importantly, personality traits are considered more accurately conceptualised as continuous rather than binary or categorical constructs [17]. Some prediction efforts estimated traits on a continuous scale for Big Five personality impressions (i.e., third-person annotations) of YouTube vlogger videos using LIWC [18]. Conscientiousness was best predicted (RMSE = 0.64 on a scale of 1-7, $R^2$=0.18). Another study predicted Dark Triad traits (narcissism, Machiavellianism, and psychopathy) from Twitter data including unigrams, LIWC categories, and profile picture features, with ground truth established through a survey [19]. The best model showed a correlation of $r$=0.25 between predicted and observed values [19]. Another study showed poor performance in both regression and classification tasks for Big Five and Dark Triad prediction on Twitter with LIWC measures, even though correlations between personality traits and LIWC categories were found [20].

Various other studies attempted predicting demographic variables like age and gender. In the PAN[2] shared task on this topic, best performance for predicting five age classes was 58.97% (random baseline: 0.19) using stylistic features and vector representations of terms and documents [21]. Using stylometric features and $n$-grams they correctly classified gender 75.64% of the time (compared to a random baseline of 0.56) [21]. In another effort, using unigrams to predict age on a continuous scale achieved a mean absolute error of approximately four years [22]. Furthermore, gender classification on Twitter using *n*-grams achieved 91.80% accuracy when using all tweets from a profile [23].

---

[1] Considered to be a small effect, where small = 0.2, medium = 0.5, large = 0.8 [14]

[2] Plagiarism analysis, Authorship identification, and Near-duplicate detection: https://pan.webis.de/

## 1.3 Abusive and violent language

Author profiling is also gaining traction within threat assessment where the source of an abusive, threatening, or extremist text posted online needs to be determined. The Profile Risk Assessment Tool (PRAT), which is intended for risk assessment of violent written communications, constructs a text author's personality profile [7]. The profiles are constructed through IBM Watson Personality Insights, which predicts Big Five traits with models trained on word embeddings from a large ground truth dataset. IBM Personality Insights has also been used to study texts authored by mass murderers [24]. Personality traits measured in the mass murderer texts were compared to population medians, with the former scoring higher on openness, but lower on extraversion and agreeableness [24]. In a study on profiling the texts of school shooters, personality profiles were constructed by means of word embeddings [25]. Distances were calculated between embedding representations of traits (e.g., 'narcissism' and 'paranoia') and school shooter texts ($n = 6$) as well as neutral writing ($n = 6,056$). After ranking all texts on these measures, all school shooter texts could be identified by examining 3% of the entire corpus [25].

In our view, an important step has been missed in the use of author profiling for threat assessment; the link between author characteristics and abusive language remains unestablished. Although the relationship between neutral language and author characteristics such as personality [8], age [9] and gender [10] has been extensively tested, this has yet to be done for abusive language. For instance, we do not know if highly extraverted persons use more swearwords in their abusive writing, or if men use more sexual words when insulting someone. In order to test these possibilities, the ground truth regarding the characteristics of the abusive text author are needed. Typically, the study of abusive language makes use of naturally occurring data, for example in comment sections [3], extremist forums [26], and on Twitter [27]. Since ground truth of gender, age, and personality of text writers is often lacking in such data, the texts in this study are experimentally elicited. This setting may be less natural than spontaneous abusive language. However, in order to make progress in this relatively novel area of research, it is important to obtain ground truth regarding author characteristics before author profiling can be performed on naturally occurring abusive data. To our knowledge, the feasibility of author profiling within the domain of abusive language has yet to be tested, and the current study serves to address this issue.

## 2 Method

### 2.1 Data availability

Features, code, and supplemental materials are available on the Open Science Framework: https://osf.io/ag8hu/.

### 2.2 Sample

800 participants were recruited through the crowdsourcing platform Prolific Academic. Only adult UK citizens with English as their first language were eligible. Participants who failed the attention checks[3] were excluded, resulting in a sample of 789.

### 2.3 Procedure

Participants wrote both a stream-of-consciousness (SOC) essay about current thoughts and feelings, and an abusive text directed at a politician. Each task lasted for at least three minutes and participants were encouraged to write at least 100 words. For the abusive writing task, participants rated eight UK politicians from most to least favourite, then were assigned to write about their negative thoughts and feelings about their least favourite politician. They were told they could be as insulting, abusive, and offensive as they wanted (writing examples are given below). Lastly, participants completed two personality tests and were asked for their gender and age.

### 2.4 Personality measures

The HEXACO-60 [28] measures honesty-humility, emotionality, extraversion, agreeableness versus anger, conscientiousness, and openness to experience, on a scale from 1 = strongly disagree to 5 = strongly agree, with 10 questions per trait

---

[3] Two questions asking participants to select a specific response (e.g. 'strongly disagree') to continue

(i.e. resulting in a score between 10-50 per trait). The Short Dark Triad (SD3) [29] measures Machiavellianism, narcissism, and psychopathy on a Likert scale from 1 = strongly disagree to 7 = strongly agree, with 9 questions per trait (i.e. a score of 9-63 per trait).

## 2.5 Data examples

**Stream-of-consciousness.** *I feel content and I am reasonably happy at this present moment in time. It may be a challenging few months for me and I am looking forward to the time ahead. Some times I do feel at times that things get on top of me and find it hard to get going in the morning. I think that the future is bright for me and I fight on with perseverance and determination even though I have had some setbacks. I overall feel more confident and determined than ever even though at times I doubt myself for a brief moment.*

**Abusive writing.** [POLITICIAN] you are a liar, *a cheat, an abhorrent person, your arrogance is beyond repair, you are determined to drag the country into the gutter, you are a complete shit with total disregard for women, I hope you die in regret of what you have dragged our country into, we are now the laughing stock of europe, I hope you rot, shame on you, you are possibly the worst politician that we have ever had, you deserve a long and hard punishment for what you've done, you utter prick, please rot in hell for a long long time I hope*

## 2.6 Statistical tests

We test for statistical relationships between author characteristics (personality, age, and gender) and LIWC2015 measures drawn from both types of text [9]. For the correlation between personality traits and the LIWC, we use a Bonferroni-corrected threshold of $0.05 / (89*9) = 0.000062$ for 89 LIWC categories and 9 personality traits (6 HEXACO + 3 Dark Triad). Although previous research on linguistic correlates of personality [8] did not apply such corrections, we argue this is appropriate in order to account for possible Type I errors caused by performing multiple correlation tests [30, 31].

Multivariate regression was used to assess the effect of age (and quadratic age, here: the absolute difference from age 40) on all LIWC2015 categories, while controlling for gender, following

[12]. We also assess whether there is a multivariate effect of gender in a MANOVA for all LIWC2015 categories, then perform univariate post-hoc ANOVAs, following [13].

## 2.7 Prediction tasks

All prediction and classification tasks below are performed for stream-of-consciousness and abusive writing separately. The feature sets were:

1. Number of words (baseline model)
2. Stemmed uni- and bi-grams (with stop words removed)
3. Parts-of-speech (universal POS tags from the R implementation of SpaCy [32]).
4. All 89 LIWC2015 categories. In the abusive writing condition, we also include the proportion of abusive language[4] words as feature.
5. Composite feature set: all of the above features.
6. Filtered feature set: a selection of features from the composite feature set, filtered using a General Additive Model [34], and included if there is a functional relationship ($p < 0.05$) between the feature and outcome variable, during ten resampling iterations [35].
7. Pre-trained word embeddings, using the GloVe 6B corpus (100 dimensions) [36].
8. Pre-trained BERT language model (base uncased model with 12 layers and 768 hidden nodes) which takes into account contextual relations between words through bi-directional training [37].

All tasks are performed with a 10-fold-10-repetitions cross validation on the training set (80% of the data). The remaining 20% of the sample was used as a hold-out test set. The prediction analysis included the following steps:

- Predicting the HEXACO and Dark Triad traits in isolation on a continuous scale (regression model using a Support Vector Machine algorithm). Reported performance metric: Mean Absolute Error (MAE) and Mean Absolute Percentage Error (MAPE)

---

[4] A composite measure of abusive language following Kleinberg, van der Vegt, & Gill (2020), measuring profane and racist language from various dictionaries.

- Predicting partitioned personality traits (binary classification with a Naïve Bayes algorithm). Following [38] we perform a median split on each personality trait. Reported performance metric: classification accuracy.
- Predicting author age (regression with an SVM algorithm). Metric: MAE and MAPE.
- Predicting author gender (male or female; binary classification with a Naïve Bayes classifier). Metric: classification accuracy.

## 3 Results

### 3.1 Descriptive statistics

Participants' mean age was 37 years ($SD$ = 12.73; 63.75% male). The average word count for SOC writing was 120.51 words, and 120.62 for abusive writing, with no significant order effect found for word count. We observed differences between SOC and abusive writing (i.e., manipulation check) on 60 out of 89 LIWC categories (adjusted $p$-value of 0.05/89 LIWC categories). Furthermore, the average number of abusive words[1] in abusive writing was 4.03, with a mean of 2.05 in stream-of-consciousness writing, representing a difference of $t(788)$=16.992, $p$<0.001, Cohen's $d$=0.60. The order in which participants wrote texts did not affect the number of abusive words written in the abusive text, $t(781.88)$=-1.67, $p$>0.05. Participants who wrote the SOC essay after the abusive text, used somewhat more abusive words, $t(745.86)$=4.12, $p$<0.001, albeit with a small effect size $d$=0.29.

### 3.2 Personality

**Correlations.** Table 1 presents significant correlations ($p$ < 0.000062) between HEXACO and Dark Triad traits with LIWC2015 variables. Note that no significant correlations were found for honesty, agreeableness, conscientiousness, narcissism, and Machiavellianism with any of the LIWC variables and in neither of the writing conditions.

In short, for stream-of-consciousness writing we found significant relationships for three out of nine personality traits, and 11 out of 89 linguistic categories. For abusive writing, we saw effects for three out of nine traits and 7 out of 89 LIWC categories. The effects ranged between $r$=-0.16 to $r$=0.20 for stream-of-consciousness writing, and $r$=-0.15 and $r$=0.18 for abusive writing.

Table 1. Significant correlations LIWC and personality traits

| SOC writing | | |
|---|---|---|
| Trait | Measure | $r$ ($R^2$) |
| Emotio-nality | personal pronouns | 0.19 (0.04) |
| | first person singular | 0.20 (0.04) |
| | negative emotion | 0.14 (0.02) |
| | Anxiety | 0.18 (0.03) |
| Conscien-tiousness | word count | 0.12 (0.01) |
| Openness | word count | 0.17 (0.03) |
| | Commas | 0.19 (0.04) |
| Extra-version | Tone | 0.15 (0.02) |
| | negation | -0.15 (0.02) |
| | cognitive processes | -0.16 (0.03) |
| | differentiation | -0.16 (0.03) |
| | seeing | 0.14 (0.02) |
| | leisure | 0.15 (0.02) |
| Abusive writing | | |
| Emotio-nality | function words | 0.15 (0.02) |
| | pronouns | 0.17 (0.03) |
| | verbs | 0.15 (0.02) |
| Openness | word count | 0.20 (0.04) |
| | verbs | -0.15 (0.02) |
| | cognitive processes | -0.15 (0.02) |
| | comma | 0.18 (0.03) |
| Psycho-pathy | sexual words | 0.15 (0.02) |

**Continuous prediction.** On average, honesty, emotionality, extraversion, agreeableness, conscientiousness and openness (i.e., HEXACO traits) were predicted with an average error margin of 5.79 (MAPE = 17.2%) for SOC writing, and 5.71 for abusive writing (MAPE = 16.9%), on a scale from 10-50. The lowest error (MAE = 4.58, MAPE = 15.1%) in SOC writing was observed for predicting conscientiousness using the filtered feature set. For abusive writing this was the case for conscientiousness using parts-of-speech (MAE = 4.47, MAPE = 14.7%).

For Dark Triad predictions, the average error rate was 7.11 (MAPE = 23.6%) for SOC writing and 6.99 (MAPE = 23.2%) for abusive writing, on a scale from 1-63. The best performance in both SOC and abusive writing was obtained for psychopathy, using parts-of-speech features (MAE = 6.16, MAPE = 29.7%) and word embeddings (MAE = 5.93, MAPE = 27.7%), respectively. Importantly, a baseline model using only number of words often outperformed other feature sets. In both conditions, $n$-grams, LIWC, the composite feature set, and the BERT language model did not perform best for any of the traits.

**Classification.** We also performed binary classifications for each personality trait (based on median splits on each trait), using the same features. In SOC writing, the highest accuracy (0.63) was achieved for predicting openness (random baseline = 0.50) using BERT. For abusive writing, the highest accuracies (0.62) were achieved in predicting openness using either word embeddings or all features. The baseline feature set was never the top performer in either prediction task.

### 3.3 Age

First, we tested for possible statistical relationships between age and LIWC categories. In both writing conditions, no significant effect of age or quadratic age (while controlling for gender) on any of the LIWC2015 categories was found (all $p > 0.00056$, alpha-level adjusted for number of LIWC categories).

For the prediction of age in SOC writing, the best performing model using the filtered feature set achieved a MAE of 9.11 (MAPE = 24.61%) years. For abusive writing, best performance was achieved using word embeddings as features achieving a MAE of 10.01 years (MAPE = 27.04%).

### 3.4 Gender

We observed a significant multivariate effect of gender on LIWC2015 variables in SOC writing, Pillai's Trace = 0.30, $F(178, 1398)=1.37$, $p<0.001$. Significant differences between genders were found in SOC writing ($p < 0.00056$), where a negative Cohen's $d$ value signifies that the category was used more by men than women, and vice versa: analytical language ($d = -0.34$), pronouns ($d = 0.27$), personal pronouns ($d = 0.30$), first person singular ($d = 0.28$), verbs ($d = 0.35$), discrepancies ($d = 0.27$), focus on the present ($d = 0.26$), and apostrophes ($d = 0.28$).

For abusive writing we also found a multivariate effect, Pillai's Trace=0.32, $F(178, 1398)=1.47$, $p<0.001$. Significant gender differences were found for analytical language ($d=-0.44$), function words ($d=0.41$), pronouns

($d=0.47$), personal pronouns ($d=0.47$), first person singular ($d=0.31$), articles ($d=-0.31$), auxiliary verbs ($d=0.33$), verbs ($d=0.51$), social words ($d=0.33$), sexual words ($d=-0.24$), present focus words ($d=0.45$), and apostrophes ($d=0.26$).

For the prediction of gender in SOC writing, the highest accuracy of 0.64 was achieved using parts-of-speech as features. For abusive writing, best performing prediction accuracy was 0.70, again using parts-of-speech. It must be noted that the proportion of males in the dataset was 0.64, therefore there is practically no improvement over a model which always predicts the majority class.

## 4   Discussion

The current paper examined the feasibility of author profiling through (abusive) language. We looked at statistical relationships between LIWC variables and authors' personality traits and demographics (age, gender), and performed prediction experiments.

### 4.1   Statistical relationships

First and foremost, some statistical relationships between abusive writing and author characteristics were observed. Language use in abusive texts were related to emotionality, openness, and psychopathy scores. We also observed gender differences in abusive texts, but no significant effect of age on the abusive writing was found. Interestingly, our results seem to confirm that neutral and abusive writing are differently related to personality traits. Of particular interest is the fact that differences in language use based on differences in psychopathy can be measured in abusive writing, but did not emerge in neutral writing. Of further interest is the fact that differential gender differences emerged in abusive writing when compared to SOC writing, with men for example using more sexual words, and women using more social words.

It is important to note that the majority of LIWC categories and personality traits did not seem to be significantly related to abusive or neutral writing. We also observed fairly low correlations with personality traits, with an average of $r=0.14$ for stream-of-consciousness writing, and $r=0.12$ for abusive writing. These values are smaller than the average correlation of $r=0.23$ found elsewhere [11], and also do not reach the average of $r=0.32$ for language-based studies in particular [39]. Results were also qualitatively different from previous research, seeing as we do not observe relationships between agreeableness and conscientiousness with any linguistic variable in either writing condition, whereas previous research does report such effects [39, 40]. These disparities largely are due to the more stringent statistical criteria applied in the current study, but it can be argued that these corrections should have been applied in previous studies in the first place.

## 4.2    Prediction tasks

On average, the continuous prediction of personality traits was approximately 15% off in both neutral and abusive writing. Baseline models (using number of words) performed surprisingly well, whereas feature sets such as *n*-grams and LIWC that showed success in previous studies [19, 41] performed poorly in both abusive and neutral writing. When personality prediction was simplified into a binary classification task, accuracy was also markedly lower than in previous research [15, 16]. It must, however, be noted that performance between writing conditions did not follow the same patterns, further suggesting there is a difference between abusive and neutral writing.

When predicting age, we observed an error margin of approximately ten years in both conditions. This stands in stark contrast with previous research, which used the same or fewer features and achieved an error of four years [22], potentially because a larger amount of data (in terms of text and participants) was available. However, approximating someone's age based on their language to plus or minus ten years may be helpful in a context where there is a wide range of possible ages.

Although we achieved an accuracy of 70% for gender classification using abusive writing, this is only marginally superior to a model which always predicts the majority class. Previous attempts achieved accuracy levels in the range of approximately 75% (with a 0.56 random baseline) to 92% (0.55 baseline) with similar feature sets as in the current work [25, 15]. Again, even though we observed gender differences for various LIWC categories in abusive and neutral writing, these effects did not seem to transfer into high prediction performance.

## 4.3    Possible explanations

There are several possible explanations for these results. First of all, our writing task involved instructed online writing, which is arguably different from handwritten stream-of-consciousness essays [8, 11] or more natural, uninstructed (abusive) social media posts on Twitter or Facebook [19, 39]. The fact that participants were instructed to write abusive text when they normally may not be inclined to do so, may have lowered the external validity of the study. On the other hand, we have attempted to carefully mimic a setting in which participants may want to produce abusive language by instructing them to write about a politician they strongly dislike. Moreover, the highly anonymous nature of our task may have enabled some participants to be even more abusive than they would be in an online setting where messages can be traced back to a user profile. Lastly, the number of words (120 on average) may have impacted on our ability to adequately predict author traits from language. Nevertheless, online writing is generally short in nature, and therefore testing the ability to make predictions on short texts seems especially relevant for applying these methods to online contexts.

## 4.4    Practical significance

Whether the error rates for personality, age, and gender obtained in this study are problematic, is a matter of perspective. One could argue that a prediction of personality within 15% of the actual value is useful if a general profile of a text author is desired. The same holds for the prediction of age and gender, if one can accept a certain degree of uncertainty. For example, if the age of the author of a violent text is completely unknown, a prediction within a range of ten years (e.g., $40 \pm 10$ years) may be useful to investigators. However, within a threat assessment or law enforcement context, decisions based on such a system may have far-reaching consequences and inaccuracies may become highly problematic. For example, an inaccurate profile (e.g., the wrong gender is predicted) may lead to the identification or arrest of an innocent individual, and vice versa, the true source of a threat may be missed.

The results of this study illustrate another important point: statistical significance does not equate to practical significance. Even though we observed significant statistical relationships between author demographics and (abusive) language, these effects do not per se translate into practically relevant predictions. Indeed, work on predicting life outcomes has raised the possibility that understanding a phenomenon through statistical inference does not automatically translate to high prediction performance [42]. Yet increasingly, research focusing on violent individuals examines and predicts author characteristics through language, for example in terrorist manifestos and extremist forums [7, 24, 25]. Often, these studies refer back to original research that has 'established' a link between language and personality [8, 11], assuming that this relationship generalises to other types of language (e.g., violent or threatening texts).

The current study was the first to test this assumption in a context of abusive language, and found that these relationships are markedly different, and of little importance in constructing accurate personality profiles. As such, our study suggests that the empirical body underpinning many studies on linguistic examinations of threats and terrorism, may be weaker than how they are portrayed. While the current study demonstrates that such predictions are currently inaccurate for abusive writing as operationalised here, further research is necessary to explore if indeed there are other conditions where predictions are more successful. One future avenue may include using non-linguistic information (e.g. social media meta-data) as additional sources of information for prediction tasks. Other author characteristics may also be considered for prediction, such as education level or language proficiency (e.g., whether English is the first language of the author). The focus on age and gender in this study is straightforward because of its relevance to (criminal) investigations, whereas personality prediction was chosen due to its increased popularity in threat assessment and offender profiling [7, 25].

All in all, regardless of which author characteristics and language features are used, it remains important to realise that these predictions are highly complex. Therefore, it is crucial to consider the limitations (i.e., error margins) of these systems before they are implemented in practice.

## 5 Conclusion

This paper tested for relationships between author personality, age and gender and the way in which texts are written, with specific attention paid to abusive texts. When we used linguistic information from the texts to predict personality, age and gender, we observed some statistically significant relationships between author demographics and psycholinguistic measures. Importantly, these effects did not result in high prediction performance. The results illustrate that statistical significance does not equate practical significance. We urge researchers and practitioners in the field of threat assessment to exercise caution in using author profiling, specifically in contexts were potentially dangerous individuals are the subject of interest.